\documentclass[letterpaper]{article} % DO NOT CHANGE THIS
\usepackage{aaai2026}  % DO NOT CHANGE THIS
\usepackage{times}  % DO NOT CHANGE THIS
\usepackage{helvet}  % DO NOT CHANGE THIS
\usepackage{courier}  % DO NOT CHANGE THIS
\usepackage[hyphens]{url}  % DO NOT CHANGE THIS
\usepackage{graphicx} % DO NOT CHANGE THIS
\usepackage{natbib}  % DO NOT CHANGE THIS AND DO NOT ADD ANY OPTIONS TO IT
\usepackage{caption} % DO NOT CHANGE THIS AND DO NOT ADD ANY OPTIONS TO IT
\usepackage{algorithm}
\usepackage{newfloat}
\usepackage{listings}
\DeclareCaptionStyle{ruled}{labelfont=normalfont,labelsep=colon,strut=off} % DO NOT CHANGE THIS
\floatstyle{ruled}
\newfloat{listing}{tb}{lst}{}
\floatname{listing}{Listing}
\newcommand{\qdel}[1]{}

\newcommand{\ddel}[1]{}

\usepackage{amsmath}
\usepackage{amsthm}
\usepackage{amsfonts}
\usepackage{booktabs}  % for /ref{sec} command
\usepackage{multirow}
\usepackage{xcolor}
\graphicspath{{figures/}}
\usepackage[table]{xcolor}  % for \rowcolor command
\usepackage{makecell}  % cross-line cell in the table
\floatname{listing}{Code}

\definecolor{codecomment}{rgb}{0,0.45,0.45}
\definecolor{codekeyword}{rgb}{0,0,0.65}
\definecolor{codestring}{rgb}{0.6,0.3,0}
\definecolor{codelinenum}{gray}{0.4}
\lstdefinestyle{pycode-style}{
  language=Python,
  basicstyle=\ttfamily\small,
  numbers=left,
  numberstyle=\tiny\color{codelinenum},
  keywordstyle=\color{codekeyword},
  commentstyle=\color{codecomment},
  stringstyle=\color{codestring},
  breaklines=true,
  showstringspaces=false,
  mathescape=true,
}
\lstnewenvironment{codeblock}[1][]{
  \begingroup
  \everymath{\color{codecomment}}
  \lstset{style=pycode-style,#1}
}{
  \endgroup
}
\usepackage[noend]{algpseudocode}

\allowdisplaybreaks[4]
\newtheorem{example}{Example}

\title{PrefixGPT: Prefix Adder Optimization by a Generative Pre-trained Transformer\thanks{Corresponding author: Weikang Qian.}}
\author{
    Ruogu Ding\textsuperscript{\rm 1},
    Xin Ning\textsuperscript{\rm 1},
    Ulf Schlichtmann\textsuperscript{\rm 2},
    Weikang Qian\textsuperscript{\rm 1}
}
\affiliations{
    \textsuperscript{\rm 1}Global College, Shanghai Jiao Tong University, China\\
    \textsuperscript{\rm 2}Chair of Electronic Design Automation, Technical University of Munich, Germany\\
    Emails: rg.ding@sjtu.edu.cn, xin.ning1412@sjtu.edu.cn, ulf.schlichtmann@tum.de, qianwk@sjtu.edu.cn
}

\begin{document}

\maketitle

\begin{abstract}
Prefix adders are widely used in compute-intensive applications for their high speed.
However, designing optimized prefix adders is challenging due to strict design rules and an exponentially large design space. 
We introduce PrefixGPT, a generative pre-trained Transformer (GPT) that directly generates optimized prefix adders from scratch.
Our approach represents an adder's topology as a two-dimensional coordinate sequence and applies a legality mask during generation, ensuring every design is valid by construction.
PrefixGPT features a customized decoder-only Transformer architecture. The model is first pre-trained on a corpus of randomly synthesized valid prefix adders to learn design rules and then fine-tuned to navigate the design space for optimized design quality. 
Compared with existing works, PrefixGPT not only finds a new optimal design with a 7.7\% improved area-delay product (ADP) but exhibits superior exploration quality, lowering the average ADP by up to 79.1\%.
This demonstrates the potential of GPT-style models to first master complex hardware design principles and then apply them for more efficient design optimization.
\end{abstract}

% Uncomment the following to link to your code, datasets, an extended version or similar.
% You must keep this block between (not within) the abstract and the main body of the paper.
\begin{links}
    \link{Code}{https://github.com/Mightlaus/PrefixGPT-AAAI26}
    % \link{Datasets}{https://aaai.org/example/datasets}
    % \link{Extended version}{https://arxiv.org/abs/2408.xxxxx}
\end{links}

\section{Introduction}

Adders are important arithmetic units used in many applications such as signal processing and artificial intelligence (AI).
Among many available designs, prefix adders stand out for their fast speed~\cite{ladner1980parallel-prefix}.
They are typically modeled as prefix graphs with rigid design rules, and their design space is exponentially large. 
Consequently, the central goal of prefix adder optimization is to navigate this space to find valid designs with an optimized trade-off between circuit area and delay.

Historically, classic designs like the Sklansky~\cite{sklansky1960conditional}, Kogge-Stone~\cite{kogge1973parallel}, and Brent-Kung~\cite{brent1982regular} adders were discovered manually.
To explore the design space more systematically, design automation methods were developed.
Some heuristic approaches, such as simulated annealing~\cite{moto2018prefixsequence} and space pruning~\cite{ma2018cross}, often struggled to escape suboptimal regions.
Recently, AI-based methods using reinforcement learning (RL)~\cite{roy2021prefixrl}, gradient descent~\cite{song2024circuitvae}, Monte Carlo tree search (MCTS)~\cite{NEURIPS2024_arithmatictree}, and large language models (LLMs)~\cite{xiao2024prefixllm} have shown greater success.

\begin{figure}
  \centering
  \includegraphics[width=1\linewidth]{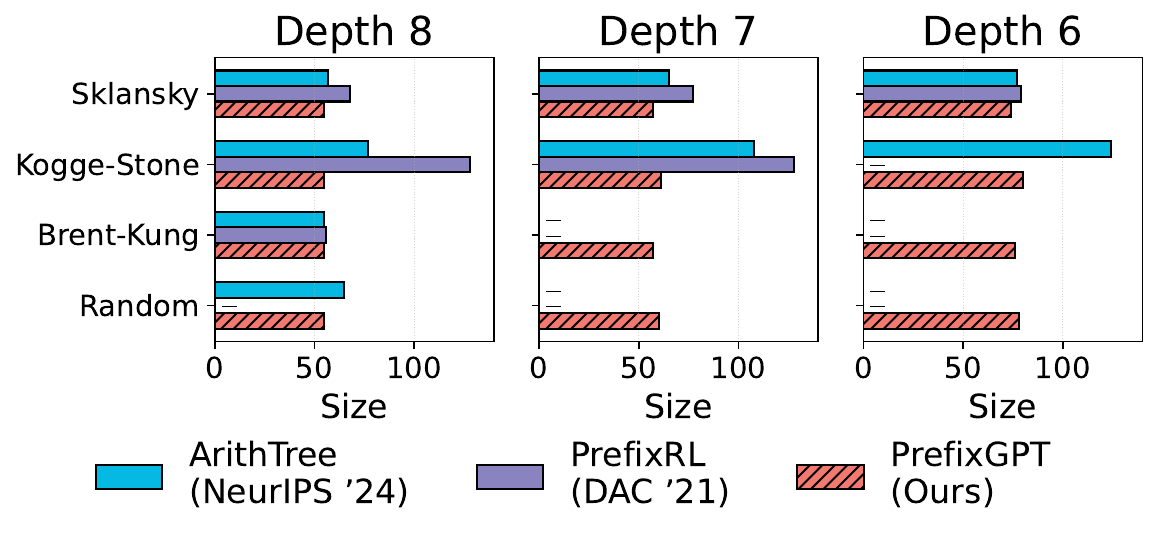}
  \caption{Optimized 32-bit adders discovered by two state-of-the-art methods and PrefixGPT under four initializations (Sklansky, Kogge-Stone, Brent-Kung and Random). Here, depth and size approximate circuit delay and area, respectively; the smaller the better. While the other methods fluctuated or even failed to find a solution (denoted by '-'), PrefixGPT was robust and consistently found better designs across all initializations. A full comparison is in Table~\ref{table:topology-analysis}.}
  \label{fig:intro-comp}
\end{figure}

Despite their achievement, these AI methods often share a common limitation: they frame optimization as a structure refinement task.
Whether by adding or deleting nodes~\cite{roy2021prefixrl,NEURIPS2024_arithmatictree} or editing textual representations of local connections~\cite{xiao2024prefixllm}, they start with an existing design and progressively modify it.
This paradigm introduces two issues.
First, iterative modifications frequently violate the strict design rules of prefix adders. 
This necessitates an additional and often inconvenient repair step to ensure correctness, which may also introduce extra hardware cost. 
Second, the performance is highly sensitive to the initial design (e.g., Sklansky).
As illustrated in Fig.~\ref{fig:intro-comp}, the effectiveness of two representative methods, PrefixRL~\cite{roy2021prefixrl} and ArithTree~\cite{NEURIPS2024_arithmatictree}, can dramatically change when starting from different manual designs.
This reveals a strong \emph{initialization bias} that may prevent these refinement-based methods from discovering high-quality designs.

The recent success of generative pre-trained Transformer (GPT)-style models offers a new perspective~\cite{openai2018improving,touvron2023llama}.
They have showed an exceptional ability in mastering complex natural language
and producing high-quality text that simultaneously satisfies structural and logical constraints~\cite{radford2019gpt2,guo2025deepseekr1}.
Prefix adders, with their rigid construction rules, form an even more structured domain than natural language.
This motivates our central question:
\textit{Can a GPT-style model also be developed to learn the ``grammar'' of prefix adders and directly generate high-quality designs that are valid by construction?}

We affirmatively answer the above question by proposing PrefixGPT, a customized GPT-style model trained from scratch for generating optimized prefix adders. 
We formulate the prefix adder design as a sequence generation problem and build a GPT-style model to generate the sequence. We recast the design rules into a dynamic legality mask that is applied during generation, ensuring every adder is valid by construction to avoid any repairs. 
After pre-trained on a large corpus of randomly synthesized prefix adders to learn the design rules, the model is further fine-tuned with an RL objective to generate prefix adders with optimized area and delay. 
Experiments show that PrefixGPT finds superior designs quickly and robustly.
Even under random initialization, its performance outperforms state-of-the-art (SOTA) methods that rely on good initial manual designs, as shown in Fig.~\ref{fig:intro-comp}.
Our contributions are summarized as follows:
\begin{itemize}
    \item We introduce a sequence representation for prefix adders, thereby reformulating their optimization as a sequence generation task that can be addressed by powerful GPT-style models.
    \item We recast the design rules for prefix adders into a dynamic legality mask that is directly integrated into the generation process, guaranteeing every design is valid by construction and eliminating any repair steps.
    \item We build PrefixGPT and equip it with a comprehensive pre-training and fine-tuning pipeline, enabling the model to leverage the learned design rules for efficient and robust prefix adder optimization.
    \end{itemize}

\section{Preliminaries}

\subsection{Prefix Adder} \label{sec:prelim-prefix}
Adders are fundamental arithmetic units in digital systems. 
Given two $n$-bit binary inputs, $A = (a_{n-1} a_{n-2} \ldots a_0)_2$ and $B = (b_{n-1} b_{n-2} \ldots b_0)_2$, an $n$-bit adder computes an $n$-bit sum $S = (s_{n-1} s_{n-2} \ldots s_0)_2$ and a final carry-out $c_{n-1}$.
A prefix adder calculates its sum bit $s_i$ ($1 \le i \le n-1$) from a carry-out at bit position $(i-1)$, denoted as $c_{i-1}$.
Therefore, a prefix adder needs to generate 
$n$ carry-outs $c_{0}, c_{1}, \ldots, c_{n-1}$.

\begin{figure*}
  \centering
  \includegraphics[width=0.9\linewidth]{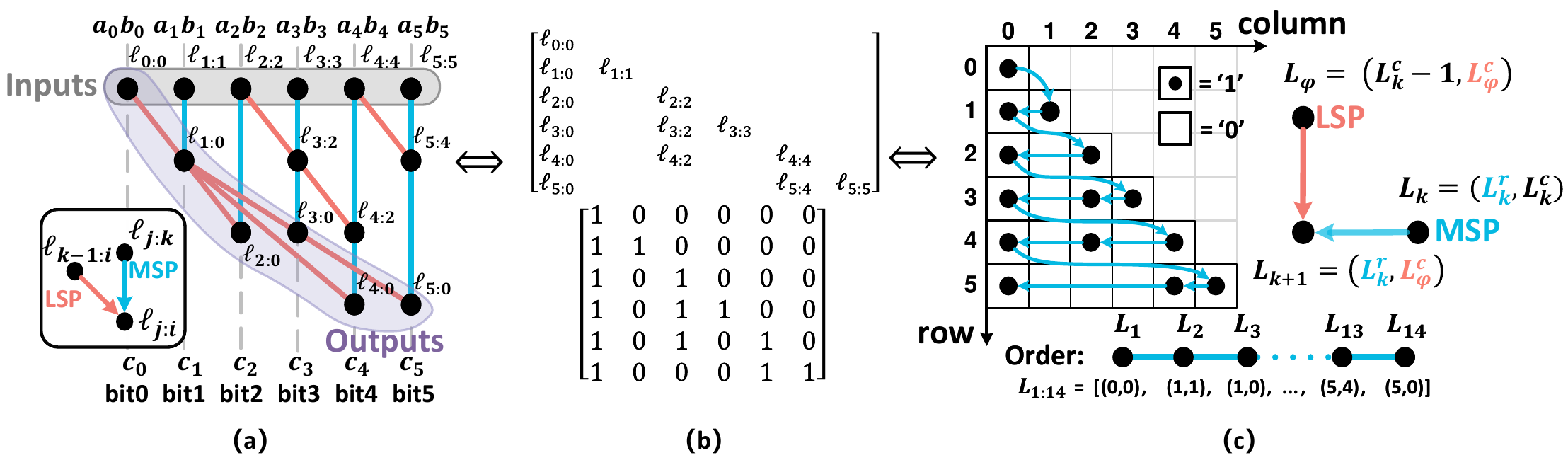}
  \caption{Equivalent representations of a 6-bit prefix adder with $\mathrm{size}=8$ and $\mathrm{depth}=4$. (a) Graph. (b) Matrix. (c) Coordinate Sequence. The blue line always connects a node and its MSP, while red line for its LSP.}
  \label{fig:prefix-reprs}
\end{figure*}

The dominant part of a prefix adder is a sub-circuit for generating these carry-outs, which can be abstracted as a \emph{prefix graph} shown in Fig.~\ref{fig:prefix-reprs}(a).
There are two types of nodes in this graph: \emph{input node}, denoted as $\ell_{i:i}$ with $0\le i \le n-1$, located at the top row of the graph, and \emph{merge node}, denoted as $\ell_{j:i}$ with $0 \le i < j \le n-1$, located at any remaining row of the graph.
For any node $\ell_{j:i}$ ($0 \le i \le j \le n-1$) in the graph, it outputs a \emph{signal pair}.
An input node $\ell_{i:i}$ takes input bits $a_i$ and $b_i$ as its inputs, while a merge node $\ell_{j:i}$ takes two signal pairs from $\ell_{j:k}$ and $\ell_{k-1:i}$ as inputs, where $k$ satisfies that $i < k \le j$.
For example, the merge node $\ell_{4:2}$ in Fig.~\ref{fig:prefix-reprs}(a) takes two signal pairs from $\ell_{4:4}$ and $\ell_{3:2}$ as inputs.
We call the nodes $\ell_{j:k}$ and $\ell_{k-1:i}$ the \emph{more significant parent (MSP)} and the \emph{less significant parent (LSP)}, respectively, of the node $\ell_{j:i}$.
In a prefix adder, the carry-out $c_i$ ($0 \le i \le n-1$) is a signal from the output of $\ell_{i:0}$.
In order to generate all carry-outs, the prefix graph must include $n$ nodes $\ell_{0:0}, \ell_{1:0}, \ldots, \ell_{n-1:0}$.
For readers interested in the actual circuit implementation of the input and merge nodes, we refer them to the appendix. In summary, a valid prefix graph should satisfy the following three design rules:
\begin{enumerate}
\item \textbf{Input rule}: It includes nodes $\ell_{0:0}, \ell_{1:1}, \ldots, \ell_{n-1:n-1}$.
\item \textbf{Output rule}: It includes nodes $\ell_{0:0}, \ell_{1:0}, \ldots, \ell_{n-1:0}$.
\item \textbf{Merge rule}: For any merge node $\ell_{j:i}$ ($0 \le i < j \le n-1$) in the graph, there must exist an integer $k$ such that $i < k \le j$ and both nodes $\ell_{j:k}$ and $\ell_{k-1:i}$ exist in the graph.
\end{enumerate}

The structure of the prefix graph directly affects the area and delay of an adder.
Specifically, the \emph{size} of the graph, which is the total number of merge nodes in the graph, is proportional to the adder area, while the \emph{depth} of the graph, which is the number of levels of the graph, is proportional to the delay of the adder.
To design a prefix adder with optimized area and delay translates to finding a prefix graph with optimized size and depth.

\subsection{Generative Pre-trained Transformers (GPTs)}
GPTs have been successfully used to tackle tasks in diverse fields, including biology~\cite{luo2022biogpt}, chemistry~\cite{m2024augmenting}, finance~\cite{seshakagari2025financial}, and programming~\cite{bucaioni2024programming}.
These GPT-style models are typically built upon a decoder-only architecture~\cite{openai2018improving} and operate \emph{autoregressively}: given a sequence of input tokens $(x_1,\dots,x_{t})$, the model predicts the next token $x_{t+1}$.
The generated token is then appended to the sequence, forming a new input sequence $(x_1,\dots,x_t, x_{t+1})$ for the subsequent generation step.
This process repeats until a terminal condition, such as an end-of-sequence (EOS) token or a maximum length limit, is met.

A GPT-style model is typically trained in two stages.
The first stage is \emph{pre-training}, where the model is trained on a large corpus of general-domain text to build a broad understanding of syntax and factual knowledge.
The second stage is \emph{fine-tuning}, where it is adapted for a specific downstream task and often uses techniques like RL to align its behavior with the desired objectives.
We employ the same two-stage training process for our PrefixGPT.

\section{Methods}

This section describes the details of PrefixGPT.
To facilitate the development of a GPT-style model for prefix adder generation, Sec.~\ref{sec:serialization} presents a method that converts a prefix graph into a coordinate sequence.
It also shows how to recast the design rules into a dynamic legality mask used in the generation process.
A custom decoder-only Transformer model (Sec.~\ref{sec:model-arch}) with a full pre-training (Sec.~\ref{sec:pre-train}) and fine-tuning (Sec.~\ref{sec:fine-tune_and_generate}) pipeline is then proposed for optimized adder exploration.
% Sec.~\ref{sec:model-arch} then describes the model architecture that builds upon the sequence representation to enable prefix adder generation.
% Sec.~\ref{sec:pre-train} details the pre-training strategy, and Sec.~\ref{sec:fine-tune_and_generate} presents the fine-tuning and generation process.

\subsection{Sequence Representation of Prefix Adder} \label{sec:serialization}
To develop a GPT-style model for generating prefix adders, we need to find a suitable sequence representation of the underlying prefix graph.
We first convert the graph into a matrix representation proposed in~\cite{roy2021prefixrl}.
As shown in Fig.~\ref{fig:prefix-reprs}(b), the prefix graph of an $n$-bit prefix adder can be represented as an $n \times n$ binary lower-triangular matrix, where the entry at row $j$ and column $i$ ($0 \le i \le j \le n-1$) is a `$1$' if and only there exists a node $\ell_{j:i}$ in the prefix graph.
We call the matrix \emph{prefix matrix}.
We record the coordinates of the `$1$'s by scanning the prefix matrix from the top row to the bottom, and within each row, from the diagonal entry to the leftmost.
This scan creates a \textit{coordinate sequence} $L_{1:N} = (L_1, \ldots, L_N)$, where $N$ is the sequence length and $L_p = (L_p^r,L_p^c)$ ($1 \le p \le N$) is the coordinate in the matrix of the $p$-th `$1$' in the scan order with $L_p^r$ giving the row coordinate and $L_p^c$ giving the column coordinate.
Note that a coordinate $L_p=(L_p^r,L_p^c)$ in the sequence corresponds to a node $\ell_{L_p^r:L_p^c}$ in the prefix graph.

For a GPT-style model, it repeatedly generates the next sequence entry given the current partial sequence until a complete sequence is generated.
Thus, to develop a GPT-style model generating the coordinate sequence, it is important to identify the valid choices for the next coordinate $L_{k+1}$ given the current sequence $L_{1:k}$.
We distinguish two cases.

% \begin{enumerate}
% \item
\textbf{Case 1}: the last coordinate in the current sequence, i.e., $L_k$, is at column $0$ of the prefix matrix.
By the input rule described in Sec.~\ref{sec:prelim-prefix}, all diagonal entries of the prefix matrix are `$1$'.
Thus, in this case, by the scanning order, the next `$1$' we visit is located at the diagonal entry of the next row of $L_k$ (see Fig.~\ref{fig:prefix-reprs}(c)).
Thus, we have $L_{k+1} = (L_k^r+1, L_k^r+1)$.

% \item 
\textbf{Case 2}: $L_k$ is not at column 0 of the prefix matrix. Then, the scan of the current row is not finished until it reaches the `$1$' guaranteed to exist at column 0 by the output rule (Sec.~\ref{sec:prelim-prefix}).
Consequently, the next coordinate $L_{k+1}$ must be in the same row and we have $L_{k+1}^r=L_k^r$, as shown in Fig.~\ref{fig:prefix-reprs}(c).
Moreover, as $L_k$ is not at column 0 of the prefix matrix, $L_{k+1}$ cannot be a diagonal entry.
Consequently, $L_{k+1}$ must correspond to a merge node in the prefix graph.
Following~\cite{roy2021prefixrl}, we restrict the MSP of the merge node corresponding to $L_{k+1}$ to be the node corresponding to $L_{k}$.
Note that this restriction simplifies the sequence generation task but sacrificing some optimality, and we will leave it as a future work to relax this restriction.
By the merge rule in Sec.~\ref{sec:prelim-prefix}, a merge node with its MSP at column $L_k^c$ must have its LSP located at row $(L_k^c-1)$,
which in turn implies that the next column coordinate $L_{k+1}^c$ must be the column of an existing node in row ($L_k^c-1$).
Formally, we have $L_{k+1}^r=L_k^r$, and $L_{k+1}^c \in \{L_\varphi^c |0\leq\varphi<k, L_\varphi^r=L_k^c-1 \}.$
% \end{enumerate}

\begin{example}
\label{exp:next-valid-rule}
Consider the example in Fig.~\ref{fig:prefix-reprs}(c).
If the last coordinate in the partial sequence, $L_k$, is $(1,0)$, the subsequent coordinate can only be $(L_k^r+1, L_k^r+1)=(2,2)$.
Alternatively, suppose $L_k=(3,3)$.
This node acts as the MSP of $L_{k+1}$, and we can derive that the row coordinate for the LSP of $L_{k+1}$ must be $L_k^c-1=3-1=2$.
Since there are only two nodes at row 2 of the prefix matrix in Fig.~\ref{fig:prefix-reprs}(c) and their coordinates are $(2,0)$ and $(2,2)$, the valid choices for the column coordinate of $L_{k+1}$ are $0$ and $2$.
Hence, the only valid choices for $L_{k+1}$ are $(3,0)$ and $(3,2)$.
\end{example}

The above rule enables constructing a \textit{legality mask} that dynamically filters out all invalid choices during each coordinate generation step.
Specifically, when sampling the next coordinate $L_{k+1}$, the mask pre-computes all infeasible options conditioned on $L_{1:k}$ and set their probabilities to zero, thereby preventing invalid sampling.
Notably, the legality mask supports a parallel implementation on GPUs. We detail its actual implementation with an example in the appendix.

\begin{figure*}
  \centering
  \includegraphics[width=0.95\linewidth]{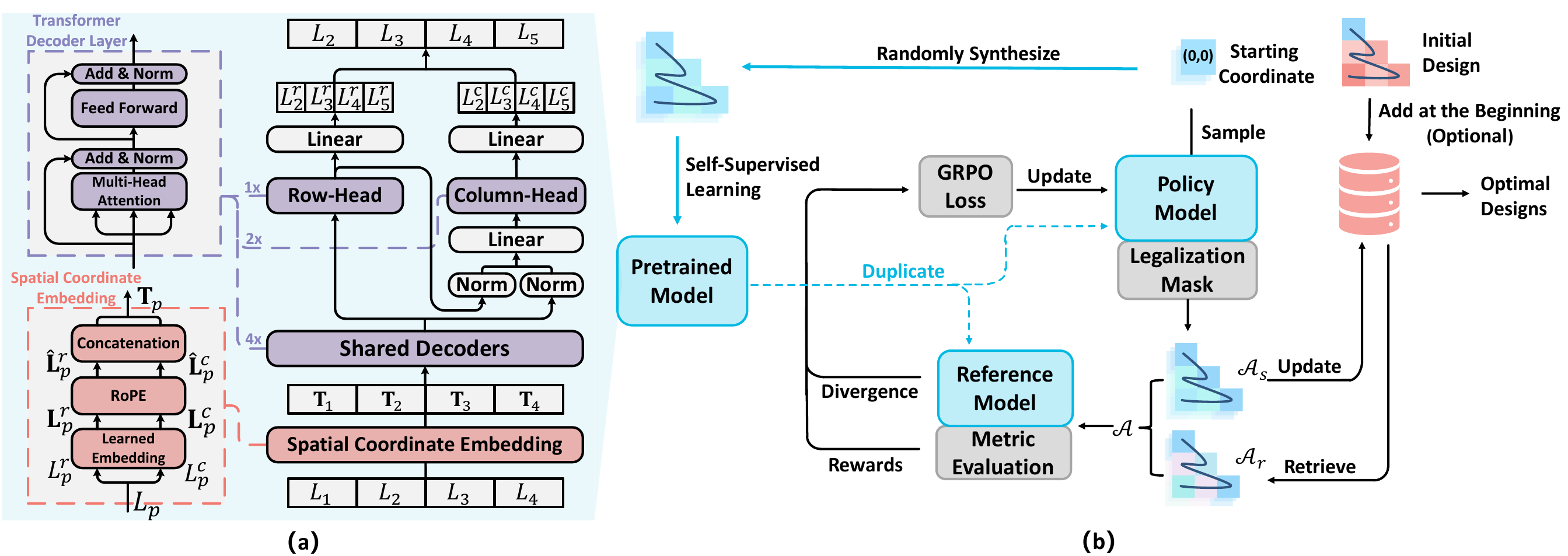}
  \caption{PrefixGPT. (a) Model architecture. (b) Training pipeline: solid blue arrows denote the pre-training phase and the solid black arrows denote fine-tuning phase.}
  \label{fig:prefixgpt}
\end{figure*}

\subsection{Model Architecture} \label{sec:model-arch}

Standard GPT-style models~\cite{radford2019gpt2} typically generate one-dimensional token sequences and lack the ability to directly handle two-dimensional (2D) coordinate data. 
To overcome this limitation, PrefixGPT introduces two techniques, a \emph{spatial coordinate embedding} and a \emph{two-head Transformer backbone}, which will be introduced next.
Fig.~\ref{fig:prefixgpt}(a) illustrates the overall architecture.

\subsubsection{Spatial Coordinate Embedding}

The spatial coordinate embedding is designed to effectively encode each node’s 2D coordinates and their relative positional information into unified tokens. 

As shown in the bottom-left box in Fig.~\ref{fig:prefixgpt}(a), we apply separate learned embedding layers to the row and column indices of each coordinate $L_p=(L_p^r, L_p^c)$, producing two $d$-dimensional vectors $\mathbf{L}_p^r$ and $\mathbf{L}_p^c \in \mathbb{R}^{d}$.
These are enhanced with the rotary positional embedding (RoPE) technique~\cite{su2024rope} to encode relative-position information.
Conceptually, RoPE associates each integer position $m$ with an orthogonal rotation matrix $\mathbf{R}(m) \in \mathbb{R}^{d\times d}$, which has the key properties of a rotation: $\mathbf{R}(m)^\top \mathbf{R}(m) = \mathbf{I}$ and, crucially, $\mathbf{R}(m)^\top \mathbf{R}(n) = \mathbf{R}(n-m)$.
Applying this to our coordinate embedding $\mathbf{L}_p^{r/c} \in \mathbb{R}^{d}$, \footnote{Throughout this paper, we use the superscript $r/c$ as a notational shorthand. It indicates that an expression or operation applies independently to both its row ($r$) and column ($c$) counterparts. For example, an entity $X^{r/c}$ represents both $X^r$ and $X^c$.} the RoPE-enhanced embedding, denoted as $\mathbf{\hat{L}}_p^{r/c}$, is obtained by a simple matrix-vector product as $\mathbf{\hat{L}}_p^{r/c} = \mathbf{R}(L_p^{r/c}) \, \mathbf{L}_p^{r/c}$. 
When computing the inner product between any two RoPE-enhanced embeddings for positions $p$ and $q$, the rotation matrices combine elegantly:
\begin{equation*}
\begin{split}
    \langle \mathbf{\hat{L}}_p^{r/c}, \mathbf{\hat{L}}_q^{r/c} \rangle &= \left( \mathbf{R}(L_p^{r/c}) \, \mathbf{L}_p^{r/c} \right)^\top \left( \mathbf{R}(L_q^{r/c}) \, \mathbf{L}_q^{r/c} \right) \\
    &= (\mathbf{L}_p^{r/c})^\top \mathbf{R}(L_p^{r/c})^\top \mathbf{R}(L_q^{r/c}) \, \mathbf{L}_q^{r/c} \\
    &= (\mathbf{L}_p^{r/c})^\top \mathbf{R}(L_q^{r/c} - L_p^{r/c}) \, \mathbf{L}_q^{r/c},
\end{split}
\end{equation*}
where $\mathbf{R}(L_q^{r/c} - L_p^{r/c})$ is solely a function of the relative distance between the coordinates. 
This enables PrefixGPT to effectively capture not only the absolute positions but also the relative distances between merge operations.

Finally, for each coordinate $L_p$, we concatenate its RoPE-enhanced embeddings to form a fused\textit{ coordinate token} $\mathbf{T}_p=\mathbf{\hat{L}}_p^r \mathbin{\|} \mathbf{\hat{L}}_p^c \in \mathbb{R}^{2d}$.
The resulting token sequence $\mathbf{T}_{1:k}=(\mathbf{T}_1,\ldots,\mathbf{T}_k)$ is then passed through the two-head Transformer backbone for next coordinate generation.

\subsubsection{Two-Head Transformer Backbone}
To sequentially predict the next coordinate in the sequence, the Transformer backbone first transforms the spatial coordinate embedding into shared hidden states $\mathbf{H}_{1:k}^{\textrm{share}}$ through four stacked Transformer decoder layers, called \emph{shared decoder}. 
These shared states then feed into a \emph{row head} and a \emph{column head}, which predict the row and column coordinates, respectively.

The row head independently refines the shared hidden states to produce the hidden states of the row, $\mathbf{H}^{\textrm{row}}_{1:k}$, with a Transformer decoder layer as $\mathbf{H}^{\textrm{row}}_{1:k} = \text{RowHead}(\mathbf{H}^{\textrm{share}}_{1:k})$.
Then, a linear projection is applied to predict the probabilistic distribution of the next row coordinate as follows:
\begin{equation*}
P(L^r_{k+1} | L_{1:k}) = \text{Softmax}(\mathbf{W}^{\textrm{row}} \mathbf{h}^{\textrm{row}}_{k}),
\end{equation*}
where $\mathbf{h}^{\textrm{row}}_{k}$ is the $k$-th component of $\mathbf{H}^{\textrm{row}}_{1:k}$, and we sample the next row coordinate from the obtained distribution.

Since a valid next column coordinate depends on the row coordinate as described in Sec.~\ref{sec:serialization}, the column head integrates both $\mathbf{H}^{\textrm{share}}_{1:k}$ and $\mathbf{H}^{\textrm{row}}_{1:k}$ to generate the hidden states of the column, $\mathbf{H}^{\textrm{col}}_{1:k}$, through two Transformer decoder layers, which is further processed by a final linear layer to predict the next column coordinate as follows:
\begin{align*}
\mathbf{H}^{\textrm{col}}_{1:k} = \text{ColumnHead}\left(\mathbf{W}_p\left(\mathbf{H}^{\textrm{share*}}_{1:k} \;||\; \mathbf{H}^{\textrm{row*}}_{1:k}\right)\right), \\
P(L^c_{k+1} | L_{1:k}) = \text{Softmax}(\mathbf{W}^{\textrm{col}} \mathbf{h}^{\textrm{col}}_{k}),
\end{align*}
where $\mathbf{H}^{\textrm{share*}}_{1:k}$ and $\mathbf{H}^{\textrm{row*}}_{1:k}$ denote hidden states after the root mean square normalization~\cite{zhang2019rms-norm}, and $||$ marks concatenation. The next column coordinate is also sampled from the distribution $P(L^c_{k+1} | L_{1:k})$.

\subsection{Pre-training} \label{sec:pre-train}
We first teach the model design rules though pre-training before asking it to generate optimized designs.
As illustrated by the solid blue arrows in Fig.~\ref{fig:prefixgpt} (b), we pre-train PrefixGPT using self-supervised learning on a corpus of one million valid coordinate sequences.
Each sequence is generated by performing a random walk starting at the same first coordinate $L_1=(0, 0)$ and proceeding by randomly selecting a valid next coordinate until the EOS coordinate $(n-1, 0)$ is reached for an $n$-bit adder.

During pre-training, PrefixGPT learns to predict the next valid coordinate by minimizing the loss between its predictions and the ground-truth labels.
For an input sequence $L_{1:k}$, its loss sums the individual cross-entropy loss from both row and column as follows:
% {\small
\begin{equation*}
    \mathcal{L}_\text{pre}= -\frac 1 k \sum_{p=1}^k \left( \log P(L_p^r|L_{1:p-1}) + \log P(L_p^c|L_{1:p-1}) \right).
\end{equation*}
% }

The pre-training is both efficient and transferable.
The training corpus can be synthesized on a single GPU with roughly one minute under our implementation settings, and once PrefixGPT is pre-trained on a maximal bit-width $n$, it can be instantly adapted to any $m$-bit tasks ($m \le n$), simply by resetting the EOS coordinate from $(n-1,0)$ to $(m-1,0)$.
For this work, PrefixGPT was pre-trained on sequences up to $n=48$, due to GPU memory constraints.

\subsection{Fine-tuning for Optimized Adder Generation} \label{sec:fine-tune_and_generate}
The fine-tuning process is indicated by the solid black arrows in Fig.~\ref{fig:prefixgpt}(b).
In this stage, we employ RL to align the output of the pre-trained model with our design objective, i.e., an optimized area-delay trade-off, and continuously exploring the solution space to discover higher-quality adders.

Fine-tuning begins by duplicating the pre-trained model into two copies: a trainable policy model $\pi_\theta$ and a frozen reference model $\pi_{\textrm{ref}}$. 
The policy model evolves to discover new designs, and the reference model preserves the pre-training behavior to avoid performance catastrophic collapse~\cite{schulman2017ppo, stiennon2020learning}.
At each RL iteration, $\pi_\theta$ samples a group of $G$ valid adders $\mathcal{A}_s=(\alpha_1,\ldots,\alpha_G)$ using the legality mask, as shown in Fig.~\ref{fig:prefixgpt}(b).
Each adder $\alpha_i$ can be represented by a coordinate sequence $L_{1:N_i}$, where $N_i$ is its length. 
We evaluate each $\alpha_i$ with a scalar reward set as the negative area-delay product (ADP) $r_i = -\text{area}(\alpha_i)\times \text{delay}(\alpha_i)$. We choose this reward since a commonly used target for circuit optimization is minimizing the ADP.
In parallel, $\pi_{\textrm{ref}}$ re-scores the same sequences to compute a Kullback-Leibler (KL) divergence $\mathbb{D}_{\textrm{KL}}$~\cite{schulman2017ppo} between the policy model and the reference model. 
We extend its approximate form for both the row and the column factors:
\begin{align*}
    \mathbb{D}_{\textrm{KL}}(L_p) &= \mathbb{D}_{\textrm{KL}}^r(L_p) + \mathbb{D}_{\textrm{KL}}^c(L_p),\\
    \mathbb{D}_{\textrm{KL}}^{r/c}(L_p) &= \frac{\pi_\theta^{r/c}(L_p)}{\pi_{\textrm{ref}}^{r/c}(L_p)}-\log\frac{\pi_\theta^{r/c}(L_p)}{\pi_{\textrm{ref}}^{r/c}(L_p)}-1,\\
    \pi_\theta^{r/c}(L_p) &= P_\theta(L_p^{r/c}|L_{1:p-1}),\\
    \pi_\textrm{ref}^{r/c}(L_p) &= P_\textrm{ref}(L_p^{r/c}|L_{1:p-1}),
\end{align*}
where $P_\theta$ and $P_\text{ref}$ represent the prediction probability given by the policy model and reference model, respectively.

We adopt the lightweight group relative policy optimization (GRPO)~ \cite{shao2024deepseekmath} method for efficient training.
Compared with proximal policy optimization~\cite{schulman2017ppo}, GRPO prevents the expensive advantage model by directly calculating the advantage $\hat{A}_i$ with a normalized reward as $\hat{A}_i=\frac{r_i-\mu_\mathcal{A}}{\sigma_\mathcal{A}}$, where $\mu_\mathcal{A}$ and $\sigma_\mathcal{A}$ are the mean and the standard deviation, respectively, of the rewards of all adders in a sample group $\mathcal{A}$. We optimize the policy model by maximizing the following objective function:
\begin{equation}
\label{eqn:RL-obj}
    \mathcal{J}(\theta) = \frac{1}{G}\sum_{i=1}^{G}\frac{1}{N_i} \sum_{p=1}^{{N_i}} \left\{\gamma^p s_\theta (L_{p})\hat{A}_i - \beta \mathbb{D}_{KL}(L_p) \right\},
\end{equation}
where $s_\theta(L_p)=\pi_\theta^r(L_p) + \pi_\theta^c(L_p)$, and $\gamma=0.99$ and $\beta=0.001$ are constants.

We also propose a \textit{best-design retrieval} mechanism to reuse advantageous designs and reinforce high-quality patterns during training.
Specifically, at each iteration, all sampled designs are logged into a database.
We retrieve from the global database of all previously generated designs a small subset $\mathcal{A}_r$ of designs with the lowest ADP, where the subset size is limited to at most 10\% of the total batch size.
The retrieved designs are then merged with the newly sampled batch $\mathcal{A}_s$ to form $\mathcal{A}=\mathcal{A}_s\cup\mathcal{A}_r$ for the GRPO update.

The best-design retrieval also supports the use of good manual designs for initialization.
Specifically, we can add good manual designs into the database at the beginning of fine-tuning. Then, those retrieved with high initial quality typically dominate the GRPO advantage calculations, gently steering the policy toward learning their superior architectures. 

\section{Experiment Results}
\subsection{Experiment Setup}
We implemented PrefixGPT in PyTorch with CUDA 12.6 and ran the experiments on a single NVIDIA RTX 4090 GPU and an AMD EPYC 9374F CPU.
The model used an embedding dimension of $d=128$.
We pre-trained the model for 5 epochs on a corpus of 1,000,000 randomly synthesized valid coordinate sequences using the Adam optimizer with a learning rate of $10^{-4}$. 
For fine-tuning, we conducted RL for $200$ iterations, and we sampled $G=512$ valid designs per iteration with a temperature of $0.8$.
The legality mask was applied through all fine-tuning iterations.

We evaluated PrefixGPT on four adder bit-widths, $16$, $24$, $32$, and $48$, and comprehensively compared it with two SOTA methods, PrefixRL~\cite{roy2021prefixrl}, which uses deep Q-learning, and ArithTree~\cite{NEURIPS2024_arithmatictree}, an MCTS-based method.
To test whether PrefixGPT overcomes initialization bias, we evaluate all methods with four initializations: three manual designs, Sklansky (Sk.), Kogge-Stone (Ko.), and Brent-Kung (Br.), and a random (Ra.) setting. 
In the random setting, ArithTree and PrefixRL started searching with a randomly generated valid design.
In contrast, PrefixGPT began its search with no initial design in its database. 
To ensure a fair comparison, we evaluated each combination of bit-width and initial design by running 10 independent trials, and each trial was strictly capped at $100,000$ candidate design evaluations, resulting in a total of $1,000,000$ evaluations per combination for each method.

We also compared PrefixGPT with a generative baseline, PrefixLLM~\cite{xiao2024prefixllm}.
However, since it only scales to 16-bit adders, and its generation time per design is over 200 seconds, which is excessively long, we used the data from its authors and the original paper for comparison.

\subsection{Comparison of Prefix Graph Size}
As mentioned in Sec.~\ref{sec:prelim-prefix}, the size and depth of a prefix graph reflect the area and delay, respectively, of the corresponding prefix adder.
Given the inherent trade-off between size and depth, following~\cite{NEURIPS2024_arithmatictree}, we set three strict depth limits for each bit-width and compare the minimum achievable size under these limits.
Specifically, for each bit-width, the depth limits are $h$, $h+1$, and $h+2$, where $h$ is the minimum possible depth for an adder of the given bit-width~\cite{snir1986depthsizetradeoffs}.
As shown in Table~\ref{table:topology-analysis}, PrefixGPT achieves the best size on all 12 pairs of bit-width and depth limit, with a maximum size reduction of 59.1\% on 48-bit with Kogge-Stone initialization.
Moreover, prior methods are highly sensitive to initialization and may produce much larger or even invalid designs (indicated by `-') when tested under tight depth limits, as even a minor structural modification can easily violate the depth limit.
Additionally, for PrefixLLM, the minimum sizes obtained under Ko. initialization of 16-bit adders are 31, 28, and 24 for depths 5, 6, and 7, respectively, which falls behind PrefixGPT.

{\renewcommand{\arraystretch}{1}
\begin{table}[ht]
    \centering
    \setlength{\tabcolsep}{1pt}
    \small
    \begin{small}
    \begin{tabular}{l c c c c c c c c c c c c c}
    \toprule
    \multirow{2}{*}{\makecell{Bit-\\Width}}
      & \multirow{2}{*}{Depth}
      & \multicolumn{4}{c}{ArithTree Size}
      & \multicolumn{4}{c}{PrefixRL Size}
      & \multicolumn{4}{c}{PrefixGPT Size} \\
    \cmidrule(lr){3-6}\cmidrule(lr){7-10}\cmidrule(lr){11-14}
      & & Sk. & Ko. & Br. & Ra. & Sk. & Ko. & Br. & Ra. & Sk. & Ko. & Br. & Ra. \\
    \midrule
    \multirow{4}{*}{16-bit}
      & 5  
        & \textbf{31}  & 45       & -            & -
        & -            & -       & -            & -
        & \textbf{31}  & \textbf{31} & \textbf{31}  & 32      \\
      & 6  
        & \textbf{25}  & \textbf{25} & \textbf{25}  & \textbf{25}
        & 26           & 45      & 26           & -
        & \textbf{25}  & \textbf{25} & \textbf{25}  & \textbf{25} \\
      & 7  
        & \textbf{24}  & \textbf{24} & \textbf{24}  & \textbf{24}
        & 25           & 27      & \textbf{24}  & 30
        & \textbf{24}  & \textbf{24} & \textbf{24}  & \textbf{24} \\
    \rowcolor{gray!15}
      \multicolumn{2}{c}{Average:}
        & 27           & 31      & 25           & 25
        & 26           & 36      & 25           & 30
        & 27           & 27      & 27           & 27      \\
    \midrule
    \multirow{4}{*}{24-bit}
      & 6  
        & \textbf{45}  & 78       & -            & -
        & -            & 88      & -            & -
        & \textbf{45}  & 50      & 50           & 64      \\
      & 7  
        & 41           & 49      & -            & -
        & 51           & 83      & -            & -
        & 42           & \textbf{40} & \textbf{40} & 42      \\
      & 8  
        & \textbf{39}  & \textbf{39} & \textbf{39}  & 40
        & 48           & 76      & -            & 65
        & \textbf{39}  & \textbf{39} & \textbf{39} & 40      \\
    \rowcolor{gray!15}
      \multicolumn{2}{c}{Average:}
        & 42           & 55      & 39           & 40
        & 50           & 82      & -            & 65
        & 42           & 43      & 43           & 49      \\
    \midrule
    \multirow{4}{*}{32-bit}
      & 6  
        & 77           & 124     & -            & -
        & 79           & -       & -            & -
        & \textbf{74}  & 80      & 76           & 78      \\
      & 7  
        & 65           & 108     & -            & -
        & 77           & 128     & -            & -
        & \textbf{57}  & 61      & \textbf{57}  & 60      \\
      & 8  
        & 57           & 77      & \textbf{55}  & 65
        & 68           & 128     & 56           & -
        & \textbf{55}  & \textbf{55} & \textbf{55} & \textbf{55} \\
    \rowcolor{gray!15}
      \multicolumn{2}{c}{Average:}
        & 66           & 103     & 55           & 65
        & 75           & 128     & 56           & -
        & 62           & 65      & 63           & 64      \\
    \midrule
    \multirow{4}{*}{48-bit}
      & 7  
        & 119          & 217     & -            & -
        & -            & 224     & -            & -
        & 102          & \textbf{94} & 98        & 102     \\
      & 8  
        & 105          & 191     & -            & -
        & 123          & 222     & -            & -
        & 99           & \textbf{87} & 88        & 100     \\
      & 9  
        & 94           & 155     & -            & -
        & 120          & 207     & -            & -
        & \textbf{86}  & \textbf{86} & \textbf{86} & \textbf{86} \\
    \rowcolor{gray!15}
      \multicolumn{2}{c}{Average:}
        & 106          & 188     & -            & -
        & 122          & 218     & -            & -
        & 96           & 89      & 91           & 96      \\
    \bottomrule
    \end{tabular}
    \end{small}
    \caption{Prefix graph sizes obtained by various methods.
    A `-' indicates that the size is not found for a method and is excluded when computing the rounded average.}
    \label{table:topology-analysis}
\end{table}}

\subsection{Comparison of Area and Delay}
To align with industrial design practice, we evaluated the synthesized results by measuring their post-synthesis area and delay. 
This was done with the open-source logic synthesis tool \texttt{ABC}~\cite{berkeleyABC} using Nangate45 Open Cell Library~\cite{nangate45} as the standard cell library.

\begin{figure}[h]
    \centering
    \includegraphics[width=0.9\linewidth]{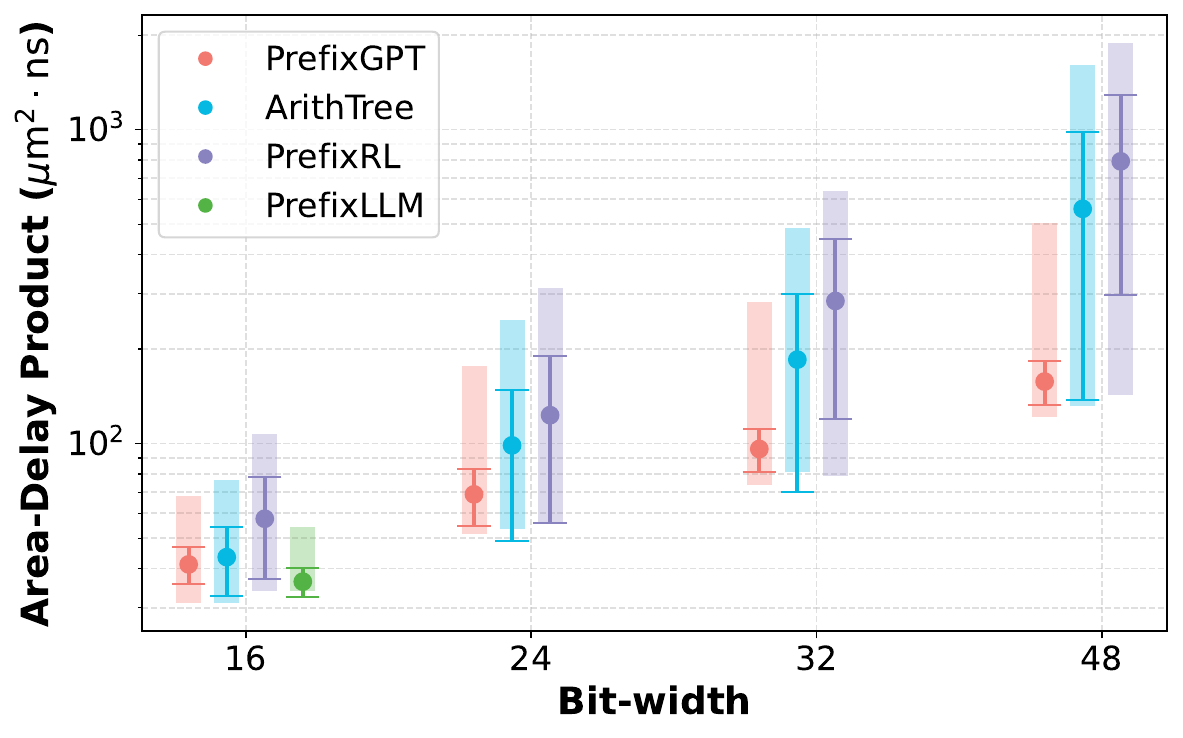}
    \caption{ADP comparison of various methods.
    In the figure, the vertical strips span the min-max range, while the markers show the mean ADP with $\pm 1$ standard deviation.}
    \label{fig:adp_comp}
\end{figure}

\begin{figure}[h]
    \centering
    \includegraphics[width=1\linewidth]{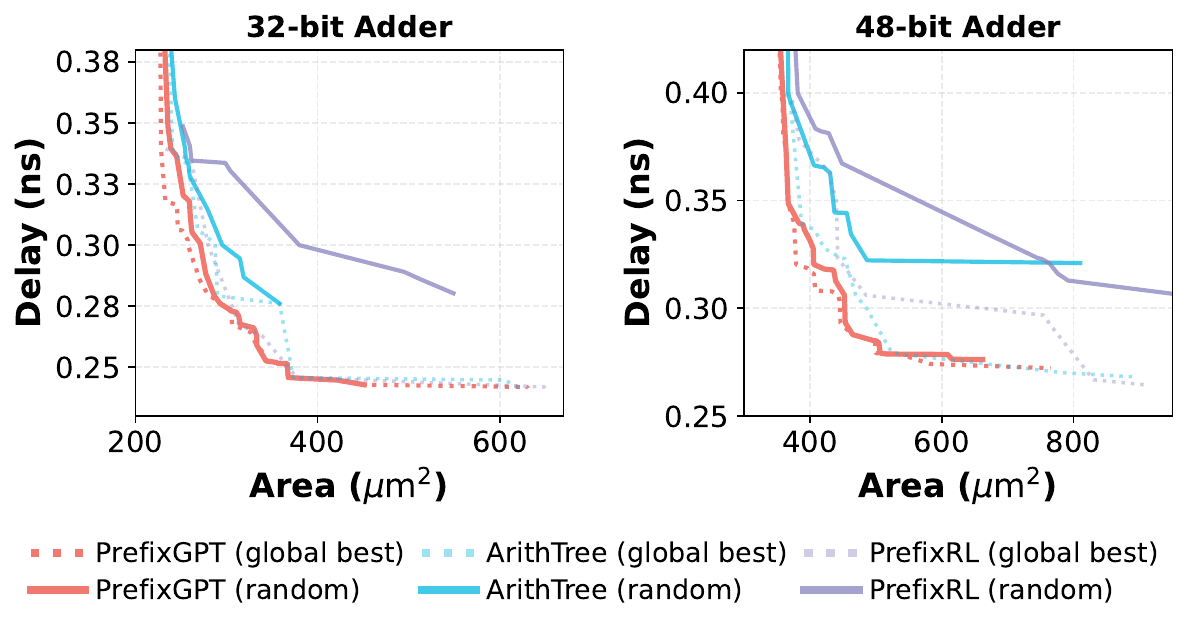}
    \caption{Prefix adder optimization results after \texttt{ABC} synthesis. In each plot, solid lines indicate the Pareto frontiers achieved from random initializations and the dashed lines represent the global Pareto frontier, aggregated from the designs across all Sk., Ko., Br. and Ra. initialization strategies.}
    \label{fig:pareto-comp}
\end{figure}

Fig.~\ref{fig:adp_comp} plots the ADP distributions, aggregated over all four initializations for each bit-width and each method.
At 16-bit, PrefixGPT achieved an minimum ADP of $31.13$ $\mu\text{m}^2\cdot\text{ns}$, better than that of PrefixLLM ( $34.01$ $\mu\text{m}^2\cdot\text{ns}$).
PrefixGPT's superiority was most evident at higher bit-widths, excelling in both performance and robustness. 
At 48-bit, it discovered a design with a minimum ADP of 121.3 $\mu\text{m}^2\cdot\text{ns}$, which was a $7.7$\% and $15.1$\% improvement over the minima found by ArithTree and PrefixRL, respectively.
Averaged over all designs, PrefixGPT achieves a mean ADP of $91.14$ $\mu\text{m}^2\cdot\text{ns}$ for 32-bit adders and $185.15$ $\mu\text{m}^2\cdot\text{ns}$ for 48-bit adders, representing a $71.9$\% and $79.1$\% reduction over the mean ADP of ArithTree.
Moreover, PrefixGPT's designs exhibit substantially greater stability. Its standard deviation in ADP is consistently lower, and at 48-bit, it shows a reduction of over 94\% compared to ArithTree ($25.2$ vs. $422.2$).

\begin{figure*}[h]
    \centering
    \includegraphics[width=0.9\linewidth]{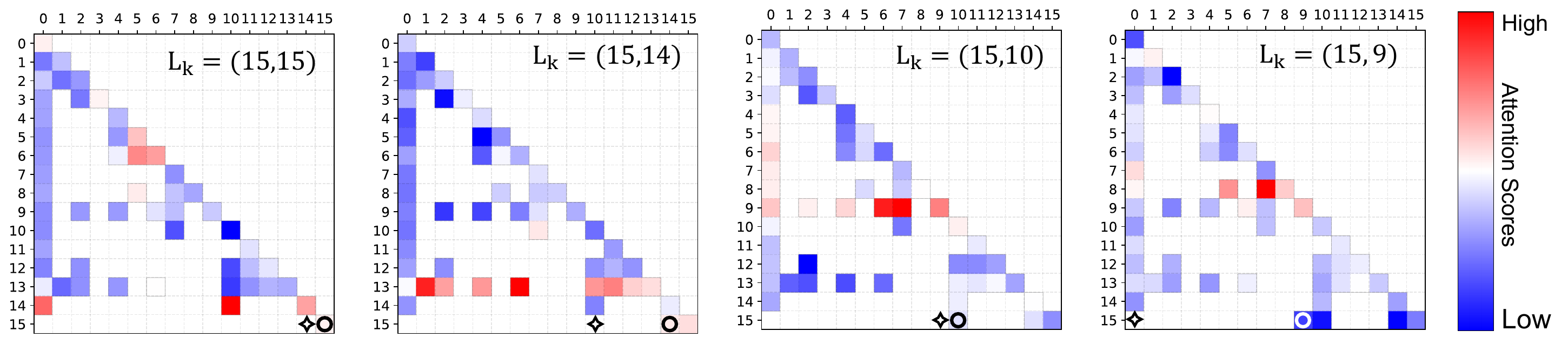}
    \caption{The model's attention scores for the final four generation steps of an unseen 16-bit adder.
    To predict the next cell, $L_{k+1}$ (starred), the model correctly focuses its attention on the potential LSPs at row $(L_k^c-1)$. Note that $L_k$ is denoted in circle.}
    \label{fig:attention_visual}
\end{figure*}

\begin{figure}
    \centering
    \includegraphics[width=0.9\linewidth]{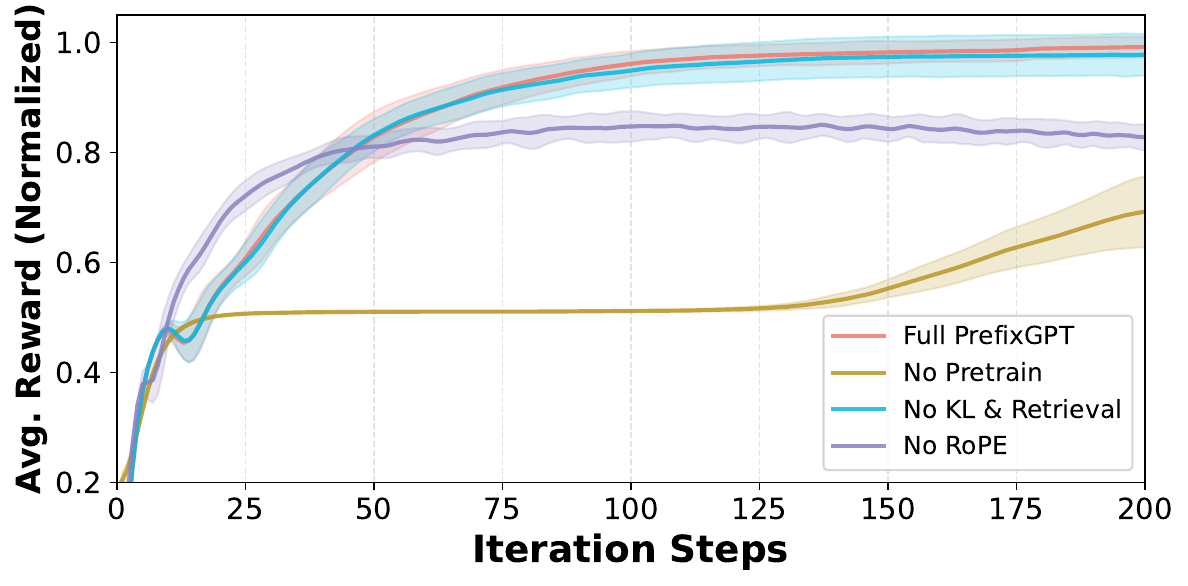}
    \caption{Ablation studies. Solid lines denote the mean average and the color stripes represent $\pm 1$ standard deviation.}
    \label{fig:ablations}
\end{figure}

This robustness is further demonstrated in Fig.~\ref{fig:pareto-comp}.
Even under random initializations (solid lines), PrefixGPT yields designs that closely match its global Pareto frontier (dashed lines), aggregated over all initialization settings.
Remarkably, these random-initialized results often outperform the best global frontiers achieved by the other methods.

PrefixGPT runs very efficiently, taking approximately 7 ms to generate one design for the 32-bit task.
We report the full wall-clock runtime in appendix.

\subsection{Investigation of the Effectiveness}
\label{sec:investigate-effect}

To investigate the sources of PrefixGPT's effectiveness, we repeated the 32-bit random optimization 20 times with different components ablated:
\begin{itemize}
    \item \textbf{In model architecture}, the RoPE block was removed from the model architecture, leaving only the learned embedding block to encode absolute positional information.
    \item \textbf{In model training}, the entire pre-training phase was omitted. Instead, a model with randomly initialized parameters was used for fine-tuning.
    \item \textbf{In fine-tuning}, the KL-divergence regularization term was removed from the objective function shown in Eq.~\eqref{eqn:RL-obj}, and the best-design retrieval was disabled. 
\end{itemize}
Fig.~\ref{fig:ablations} shows the normalized average reward over the 20 trials.
The removal of the pre-training phase and RoPE proved most detrimental, respectively resulting in 31.6\% and 16.6\% lower final average reward.
While ablating the KL-divergence and the best-design retrieval strategy led to a 1.3\% decrease in final average reward, their removal had a dramatic impact on stability, increasing the standard deviation from 0.018 to 0.038 (+110\%) at the final iteration. 

\begin{figure}[h]
  \centering
  \includegraphics[width=0.9\linewidth]{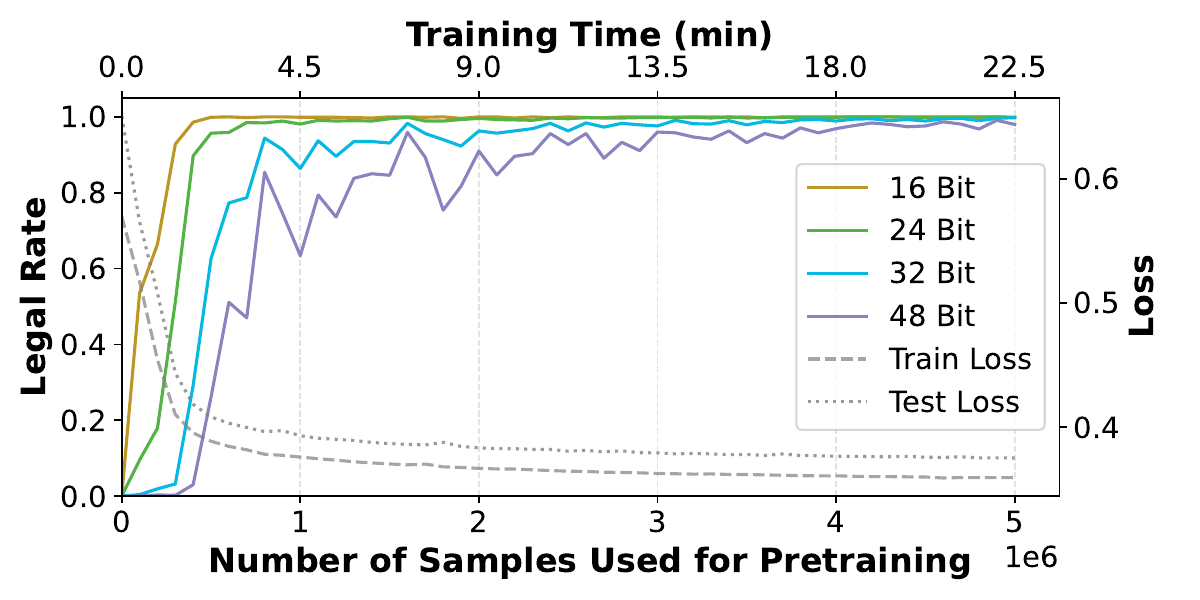}
  \caption{Generation legal rate after pre-training with varying sample sizes, evaluated without the legality mask.}
  \label{fig:pretrain-accs}
\end{figure}

These results suggest that PrefixGPT's strong performance primarily arises from its pre-training phase, where the model may internalize the fundamental design principles.
We tested this hypothesis by removing the legality mask and prompting the pre-trained model to autoregressively generate designs from the start coordinate.
As shown in Fig.~\ref{fig:pretrain-accs}, the generation legal rate (the proportion of valid designs in each generated batch) rapidly converges to nearly 100\% with sufficient training, demonstrating the model's learned ability to adhere to design rules and generate structurally valid adders.
An inspection of the model's internal mechanism provides further confirmation.
As shown in Fig.~\ref{fig:attention_visual}, the attention scores of the column head demonstrated a remarkable alignment with the merge rule described in Sec.~\ref{sec:serialization}.
For example, when $L_k=(15,14)$, the model's attention primarily focuses on coordinates in row $L_k^c-1 = 13$, which are the valid candidates for the LSP of $L_{k+1}$.
This combined evidence, from both external behavior and internal mechanism, suggests that during pre-training, PrefixGPT moves beyond pattern memorization to develop a foundational understanding of the design space, which likely enables its robust and effective search.

\section{Conclusion}
We introduce PrefixGPT, a generative pre-trained Transformer that presents a new paradigm for prefix adder optimization. 
Instead of relying on structural refinements, PrefixGPT directly generates optimized prefix-adder topologies from scratch through a sequence-based formulation. 
Extensive experiments demonstrate that our approach achieves state-of-the-art performance, discovering novel designs with improved area--delay product and substantially reduced variance across initializations.
These results show that GPT-style models can effectively handle complex hardware-design constraints and autonomously generate high-quality circuits beyond current heuristic or RL methods.

\newpage
\section*{Acknowledgments}
This work is supported by the National Key R\&D Program of China under grant number 2021ZD0114701.

\bibliography{aaai2026}

\clearpage
\appendix
\renewcommand{\thesection}{\Alph{section}} % A,B,C Index
\setcounter{section}{0} % begin from A
\onecolumn
\section{Additional Technical Details}
\subsection{Implementation of Prefix Adder}
\begin{figure}[h]
    \centering
    \includegraphics[width=0.7\linewidth]{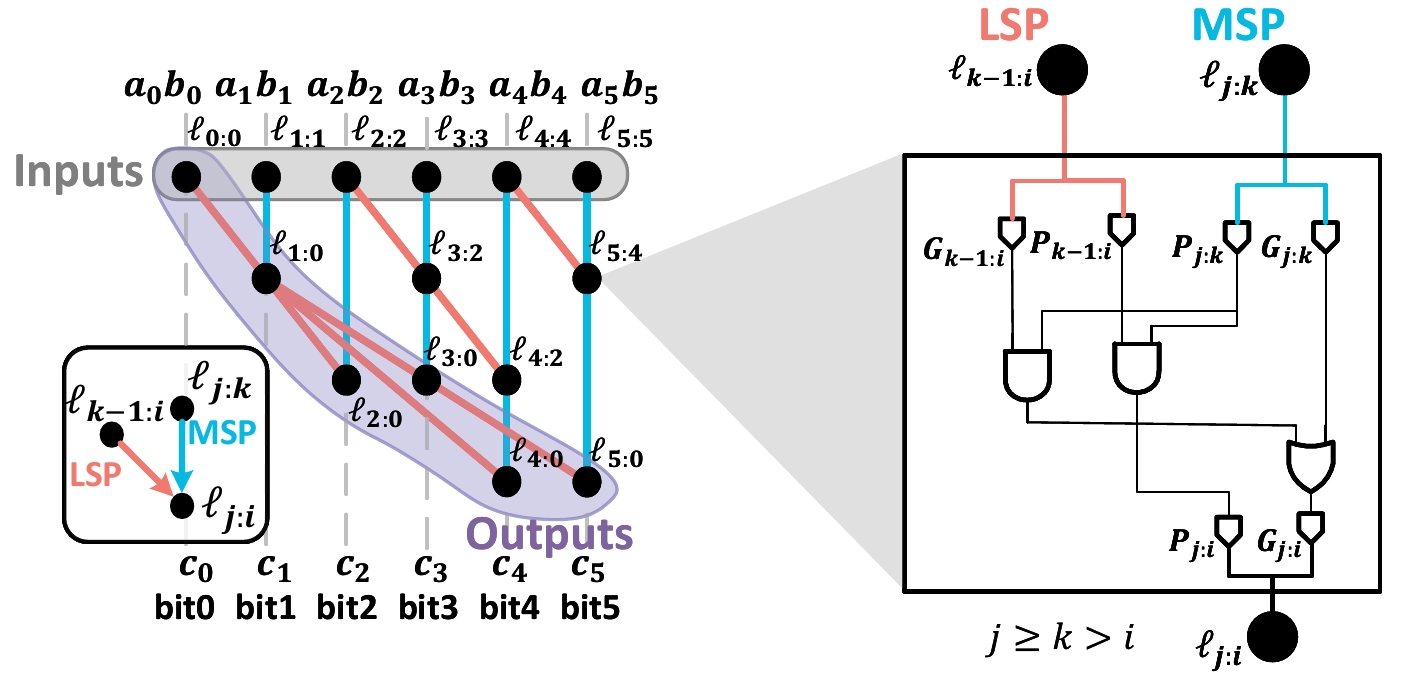}
    \caption{Gate-level implementation of a prefix adder. A merge node is implemented by a circuit} shown on the right.
    \label{fig:gate_implementation}
\end{figure}

In the implementation of an $n$-bit prefix adder, the output signal pair of each node $\ell_{j:i}$ consists of a \emph{generate signal} $G_{j:i}$ and a \emph{propagate signal} $P_{j:i}$.
They are computed as follows.
\begin{itemize}
    \item \textbf{The generate and propagate signals from an input node} $\ell_{i:i}$ ($0 \le i \le n-1$): They are directly computed from the adder's input bits $a_i$ and $b_i$ as follows:
    \begin{equation}
G_{i:i} = a_i \cdot b_i, \quad P_{i:i} = a_i \oplus b_i,
\end{equation}
where $\cdot$ and $\oplus$ denote the logic AND and XOR, respectively.
    \item \textbf{The generate and propagate signals from a merge node} $\ell_{j:i}$ $(0 \le i < j \le n-1)$: 
    They are computed by merging the signal pairs from the MSP ($\ell_{j:k}$) and LSP ($\ell_{k-1:i}$) nodes as follows:
    \begin{align}
    G_{j:i} &= G_{j:k} + (P_{j:k} \cdot G_{k-1:i}), \\
    P_{j:i} &= P_{j:k} \cdot P_{k-1:i},
    \end{align}
    where $+$ denotes logical OR. Fig.~\ref{fig:gate_implementation} shows the gate-level implementation of a merge node.
\end{itemize}

For $0 \le j \le n-1$, the carry-out at bit position $j$, $c_j$, equals $G_{j:0}$, and the sum bit at bit position $j$, $s_j$, equals $P_{j:j} \oplus c_{j-1}$.

\subsection{Implementation of Legality Mask}
We build a dynamic legality mask that follows the two cases described in Sec.~\ref{sec:serialization}.
As shown in Algorithm~\ref{alg:legal_mask_gen}, for an $n$-bit task, we compute two $n$-bit binary vectors: a row mask $\text{mask}_r$ and a column $\text{mask}_c$.

Given a partial coordinate sequence $L_{1:k}$, the entries in these masks corresponding to valid coordinates for the next step are set to $0$, while all other invalid entries are set to $1$. 
When generating the next coordinate, the model's output probability distribution is element-wise multiplied by $(1-\text{mask})$, effectively zeroing out the probabilities of all invalid positions.

\begin{algorithm}
\footnotesize
\caption{Computing legality mask.}
\label{alg:legal_mask_gen}
\begin{algorithmic}[1]
\Require Partial coordinate sequence $L_{1:k}$, bit-width $n$
\Ensure Row mask $\mathsf{mask}_r$ and column mask $\mathsf{mask}_c\!\in\!\{0,1\}^{n}$ for $L_{k+1}$

\State $(r,c)\leftarrow (L_k^r, L_k^c)$;
\Comment{Get the row and column of the last generated coordinate}

\If{$c=0$}
    \Comment{Case 1: The last node is at column 0, completing the row.}
    \State $\mathsf{mask}_r[i]\leftarrow (i\neq r\!+\!1)$;
    \Comment{The next node must be at row $r+1$.}
    \State $\mathsf{mask}_c[j]\leftarrow (j\neq r\!+\!1)$;
    \Comment{The next node must be at column $r+1$.}
\Else
    \Comment{Case 2: The last node is not at column 0; stay on the same row.}
    \State $\mathsf{mask}_r[i]\leftarrow (i\neq r)$;
    \Comment{The next node must be on the current row $r$.}
    \State $S\leftarrow\{\,L_p^c\mid p<k,\;L_p^r=c-1\,\}$;
    \Comment{Find valid LSP candidates: collect column indices of existing nodes in row $c-1$.}
    \State $\mathsf{mask}_c[j]\leftarrow (j\notin S)$;
    \Comment{Allow only columns that correspond to a valid LSP.}
\EndIf
\State \Return $(\mathsf{mask}_r,\mathsf{mask}_c)$;
\end{algorithmic}
\end{algorithm}

\begin{figure}[h]
    \centering
    \includegraphics[width=0.9\linewidth]{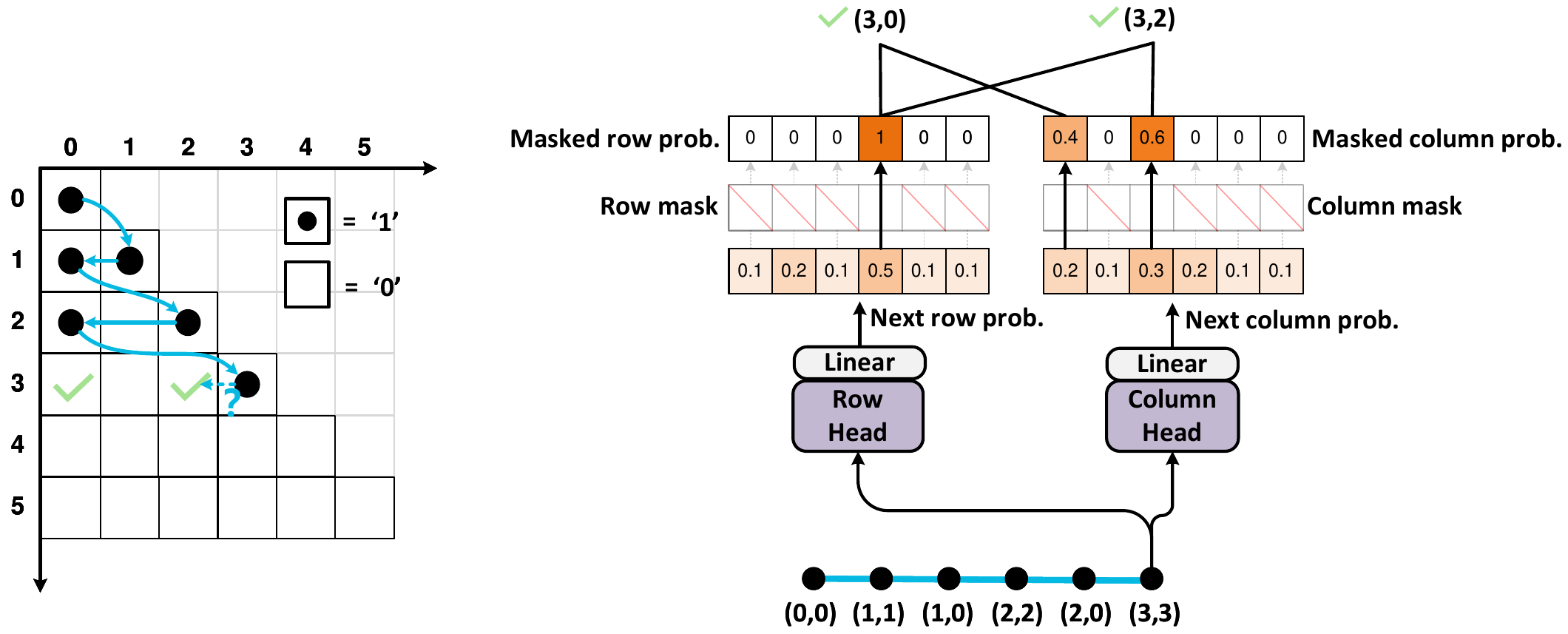}
    \caption{An example of leveraging legality mask to guarantee a valid next coordinate. Here, $L_k=(3,3)$.}
    \label{fig:mask_example}
\end{figure}

We show an example following Example~\ref{exp:next-valid-rule} from the main text:
\begin{itemize}
    \item Case 1: When the last coordinate in the sequence is $L_k=(1,0)$, the rules dictate that the next coordinate must be $(L_k^r+1, L_k^r+1)=(2,2)$.
    Consequently, both the row mask $\text{mask}_r$ and column mask $\text{mask}_c$ will have a $0$ only at index 2, with all other positions set to 1.
    \item Case 2: When $L_k=(3,3)$, the next coordinate must remain at row 3, while its column coordinate must correspond to that of an existing node at row $L_k^c-1=2$, which are $0$ and $2$, as shown in Fig.~\ref{fig:mask_example}.
    Therefore, the row mask is $[1,1,1,0,1,1]$ and the column mask is $[0,1,0,1,1,1]$, denoting all valid next coordinates are $(3,0)$ and $(3, 2)$.
\end{itemize}

To ensure reproducibility and facilitate future research, we present the directly runnable PyTorch implementation of the legality mask for parallel GPU platform in Code~\ref{lst:parallel_legal_mask}.

\section{Additional Experimental Results}

\subsection{Comparison with PrefixLLM}
\begin{figure}[!htbp]
    \centering
    \includegraphics[width=0.6\linewidth]{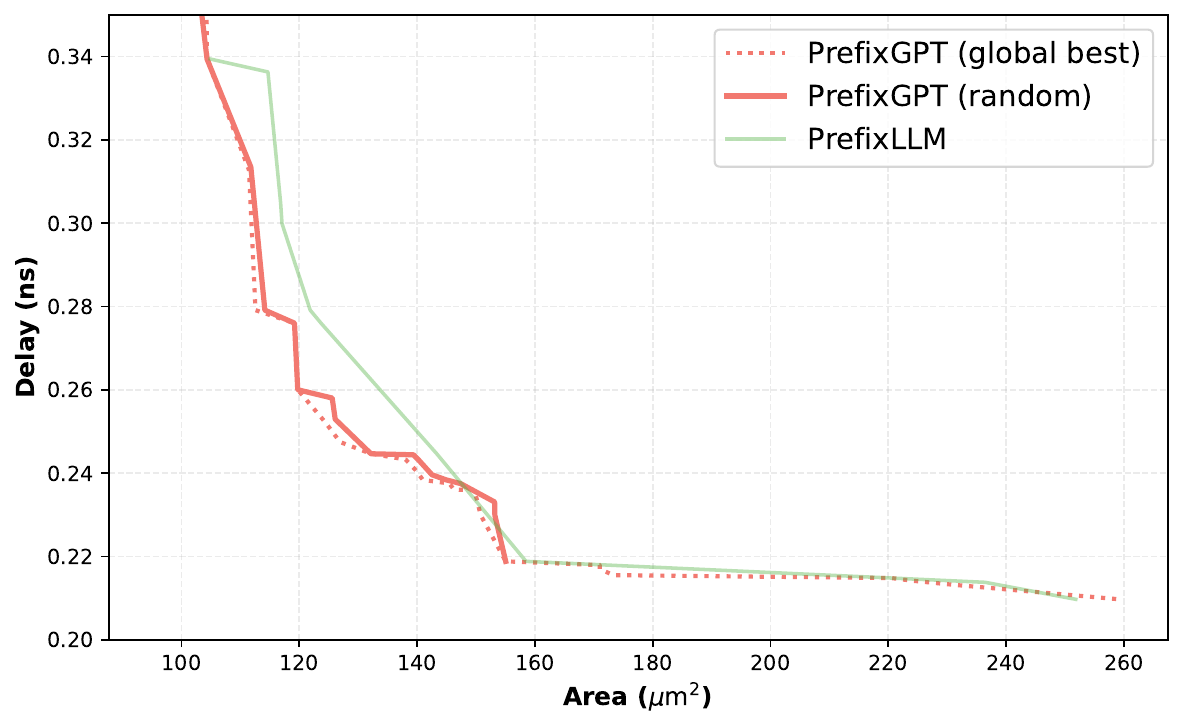}
    \caption{Area-delay Pareto frontier comparison between PrefixGPT and PrefixLLM on 16-bit adders after synthesis.}
    \label{fig:prefixllm_pareto}
\end{figure}

We collected PrefixLLM's best results under Kogge-Stone initialization on 16-bit adders and synthesized them using an identical flow as ours (i.e., using logic synthesis tool \texttt{ABC} and the Nangate45 Open Cell Library).
The area-delay Pareto frontier in Fig.~\ref{fig:prefixllm_pareto} demonstrates better performance of PrefixGPT.

\vspace{1cm}
\subsection{Wall-Clock Runtime}

Table~\ref{table:run-time} compares the average runtime among different methods.
Each value represents the time to explore 100,000 designs, averaged over various initiation strategies.
Despite generating each design from scratch, PrefixGPT exhibits a slight speed advantage on larger bits due to the batch generation process running in parallel on GPUs.

\begin{table}[!htbp]
  \centering
  \begin{tabular}{cccc}
    \toprule
    Bit-Width & ArithTree & PrefixRL & PrefixGPT \\
    \midrule
    16-bit    & 1.5       & 1.5      & 1.9        \\
    24-bit    & 8.5       & 13.3     & 5.5        \\
    32-bit    & 15.3      & 23.0     & 12.6        \\
    48-bit    & 20.3      & 53.8     & 20.5       \\
    \bottomrule
  \end{tabular}
  \caption{Average runtime (min.) for different methods.}
  \label{table:run-time}
\end{table}

\subsection{Efficiency and Scalability}
The primary bottleneck to scaling PrefixGPT for generating prefix adders with larger bit-widths lies in the VRAM usage of GPU, as GRPO depends on large-batch training for stable optimization. 
However, the growth in memory usage is linear with respect to bit-width, since the sequence length scales proportionally with bit-width as shown in Table~\ref{table:resource_usage}.
Additionally, the number of model parameters stayed constant because switching bit-widths only needs to reset the EOS coordinate (See Sec.~\ref{sec:pre-train}).

Therefore, we estimate that an 128-bit task is feasible using two 80 GB GPUs with a roughly three-hour runtime  to explore $100,000$ designs, which is sufficient to cover most practical applications. 
Additional memory-saving strategies such as gradient checkpoints, activation/KV-caching, and other lightweight optimization techniques could further reduce the VRAM requirement and extend scalability.

\begin{table}[!htbp]
    \centering
    \setlength{\tabcolsep}{4pt}
    \small
    \begin{tabular}{ccccc}
    \toprule
    \multirow{2}{*}{\shortstack{Bit-\\Width}}
      & \multirow{2}{*}{\shortstack{\# Model\\Parameters}}
      & \multirow{2}{*}{\shortstack{VRAM\\(Peak)}}
      & \multirow{2}{*}{\shortstack{Max Seq.\\Length}}
      & \multirow{2}{*}{\shortstack{Avg. Runtime\\(min.)}} \\
    & & & & \\
    \midrule
    16-bit        & 5.8\,M  & 14\,GB   & 66   & 1.9 \\
    24-bit        & 5.8\,M  & 20\,GB   & 105   & 5.5 \\
    32-bit        & 5.8\,M  & 29\,GB   & 139   & 12.6 \\
    48-bit        & 5.8\,M  & 41\,GB   & 223   & 20.5 \\
    64-bit (est.) & 5.8\,M  & $\approx$60\,GB  & $\approx$300   & $\approx$40 \\
    128-bit (est.)& 5.8\,M  & $\approx$120\,GB & $\approx$620   & $\approx$180 \\
    \bottomrule
    \end{tabular}
    \caption{Resource usage, maximum sequence length, and average runtime for PrefixGPT to generate 100k designs across different bit-widths. The 64-bit and 128-bit results are estimated.
    The maximum sequence length is empirically obtained from the generated designs, which in turn reflects the peak VRAM consumption.}
    \label{table:resource_usage}
\end{table}

\subsection{Convergence Speed}
\begin{figure}[h]
    \centering
    \includegraphics[width=0.8\linewidth]{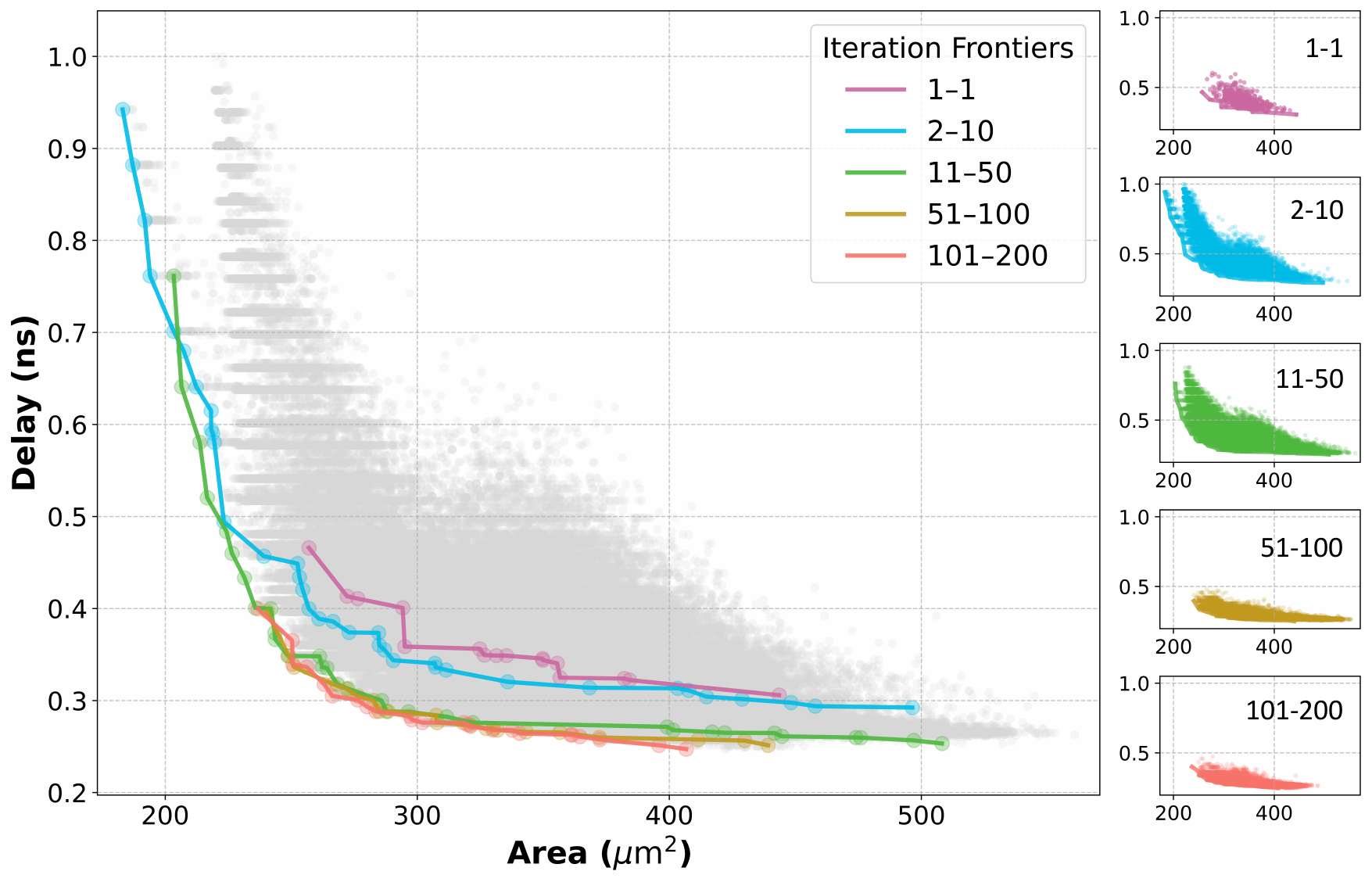}
    \caption{Convergence of PrefixGPT on a 32-bit adder with random initialization.
    The main plot (left) displays the global results aggregated over the entire fine-tuning process, while the smaller plots (right) show all designs obtained at specific iteration intervals. Each dot marks a unique design discovered during fine-tuning.
    For this task, one iteration generates 512 valid prefix adders.}
    \label{fig:convergence_frontiers}
\end{figure}

PrefixGPT converges very fast.
As shown in Fig.~\ref{fig:convergence_frontiers}, when optimizing a 32-bit adder with a random start, it approached the final Pareto frontier in fewer than 50 iterations ($<3$ min.), exploring 83.7k unique designs for 100k trials, indicating a significant sample efficiency.

\subsection{Discovered Novel Designs}
\begin{figure}
    \centering
    \includegraphics[width=0.9\linewidth]{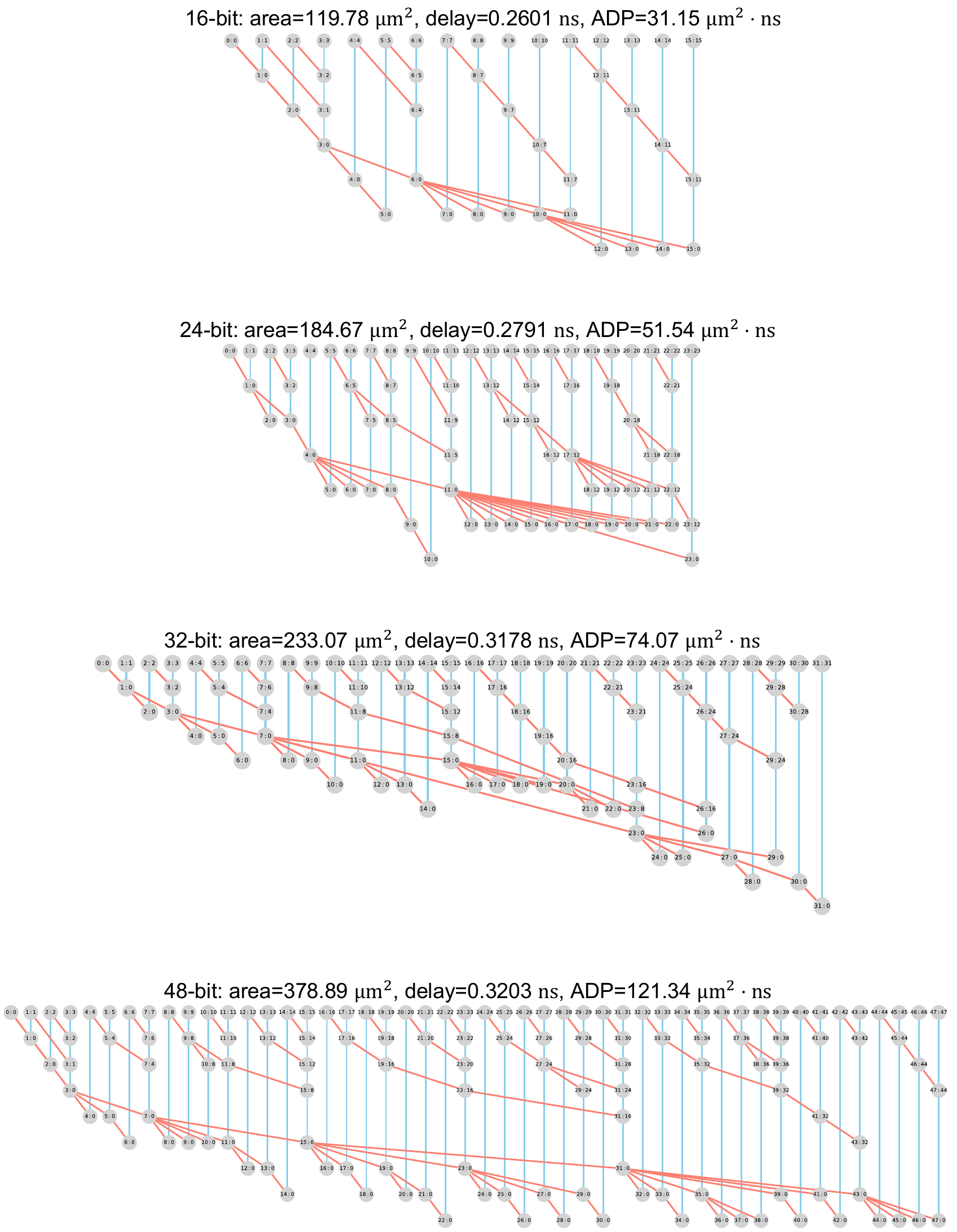}
    \caption{Discovered novel designs with the minimum ADPs for different bit-widths.}
    \label{fig:best_adp_designs}
\end{figure}

In Fig.~\ref{fig:best_adp_designs}, we present the novel prefix adder designs discovered by PrefixGPT that achieved the best performance with the minimum area-delay product (ADP).
We hope these high-quality adder architectures can not only enhance the efficiency of a wide range of digital systems, but also provide new, superior baselines for future research in electronic design automation.

\begin{listing}[h]
\caption{PyTorch implementation of the parallel legality mask.}
\label{lst:parallel_legal_mask}
\begin{codeblock}
def legal_mask(seq: torch.Tensor, n: int) -> torch.Tensor:
    # seq has shape [bs, 2, slen], bs: BatchSize, 2: (RowIdx, ColumnIdx), slen: SequenceLength
    # For example, seq[0] can be [[0,1,1,2, ...], [0,1,0,2, ...]] 
    #   representing coordinate sequence ((0,0), (1,1), (1,0), (2,2), ...).
    seq_len = torch.tensor(seq.shape[2], device=seq.device)
    seq_row, seq_col = seq[:, 0], seq[:, 1]  # [bs, slen]

    device = seq_row.device
    bs, slen = seq_row.shape
    row_indices = torch.arange(bs, device=device)  # index tensor for batch rows

    # If $L^{c}_{k}=0$, then $(L^{r}_{k+1}, L^{c}_{k+1})=(L^{r}_{k}+1,\,L^{r}_{k}+1)$. 
    case1 = (seq_col[row_indices, seq_len - 1] == 0).unsqueeze(-1).expand(-1, n + 1)
    case1_mask_row = torch.ones([bs, n + 1], dtype=torch.bool, device=device)
    case1_mask_row[row_indices, seq_row[row_indices, seq_len - 1] + 1] = False
    case1_mask_row = case1_mask_row & case1
    case1_mask_col = case1_mask_row

    # If $L^{r}_{k}\neq 0$, then $L^{r}_{k+1}=L^{r}_{k}$ and $L^{c}_{k+1}\in\{L^{c}_\varphi \mid \varphi=1,\dots,k \text{ and } L^{r}_\varphi = L^{r}_{k}-1\}$.
    case2 = ~case1
    case2_mask_row = torch.ones([bs, n + 1], dtype=torch.bool, device=device)
    case2_mask_row[row_indices, seq_row[row_indices, seq_len - 1]] = False
    case2_mask_row = case2_mask_row & case2
    # Only permit choosing LSPs from existing coordinates ($0 \leq \varphi < k$).
    time_indices = torch.arange(slen, device=device).unsqueeze(0)  # [1, slen]
    valid_steps = time_indices < seq_len.unsqueeze(-1)  # [bs, slen]
    # Compute matches where $L_\varphi^r = L_k^{r}-1$ and $0 \leq \varphi < k$.
    target_row = (seq_col[row_indices, seq_len - 1] - 1).unsqueeze(-1)  # [bs, 1]
    match_mask = (seq_row == target_row) & valid_steps  # [bs, slen]
    indices_phi = torch.nonzero(match_mask, as_tuple=True)
    col_indices = seq_col[indices_phi]
    # Case2 results
    case2_mask_col = torch.ones([bs, n + 1], dtype=torch.bool, device=device)
    case2_mask_col[indices_phi[0], col_indices] = False
    case2_mask_col = case2_mask_col & case2

    # Combine results of two cases by using logic OR
    mask_row = case1_mask_row | case2_mask_row
    mask_col = case1_mask_col | case2_mask_col

    return torch.stack([mask_row[:, :-1], mask_col[:, :-1]], dim=1)  # [bs, 2, n]
\end{codeblock}
\end{listing}

\onecolumn

\clearpage
\twocolumn

\newpage
\end{document}